\def\year{2021}\relax
\documentclass[letterpaper]{article} 
\usepackage{aaai21}  
\usepackage{times}  
\usepackage{helvet} 
\usepackage{courier}  
\usepackage[hyphens]{url}  
\usepackage{graphicx} 
\usepackage{graphicx}  
\usepackage{natbib}  
\usepackage{caption} 
\usepackage{amsfonts}

\usepackage{booktabs}
\usepackage{pifont}
\newcommand{\cmark}{\ding{51}}%
\newcommand{\xmark}{\ding{55}}%

\newcommand{\calC}{{\mathcal{C}}}
\newcommand{\calD}{{\mathcal{D}}}

\newcommand{\calL}{{\mathcal{L}}}

\newcommand{\calP}{{\mathcal{P}}}

\newcommand{\calR}{{\mathcal{R}}}

\newcommand{\calT}{{\mathcal{T}}}

\newcommand{\calW}{{\mathcal{W}}}

\newcommand{\be}{\begin{eqnarray}}
\newcommand{\ee}{\end{eqnarray}}
\newcommand{\bee}{\begin{eqnarray*}}
\newcommand{\eee}{\end{eqnarray*}}

\newcommand{\matrixb}{\left[ \begin{array}}
\newcommand{\matrixe}{\end{array} \right]}   

\newcommand{\Tref}[1]{Table~\ref{#1}}
\newcommand{\Eref}[1]{Eq.~(\ref{#1})}
\newcommand{\Fref}[1]{Fig.~\ref{#1}}

\newcommand{\Cref}[1]{Chap.~\ref{#1}}

\def\eg{\emph{e.g. }}

\def\etal{\emph{et al. }}
\def\ie{\emph{i.e. }}

\title{Optical Flow Estimation from a Single Motion-blurred Image}
\affiliations {
    KAIST Robotics and Computer Vision Lab., Daejeon, Korea \\
    dawitmureja@kaist.ac.kr, \{mibastro, rameau.fr\}@gmail.com, \{chojw, iskweon77\}@kaist.ac.kr
}
\author {

        Dawit Mureja Argaw \hspace{0.4cm}
        Junsik Kim \hspace{0.4cm}
        Francois Rameau \hspace{0.4cm}
        Jae Won Cho \hspace{0.4cm}
        In So Kweon\\
}

\begin{document}
\maketitle

\begin{abstract}
In most of computer vision applications, motion blur is regarded as an undesirable artifact. However, it has been shown that motion blur in an image may have practical interests in fundamental computer vision problems. In this work, we propose a novel framework to estimate optical flow from a single motion-blurred image in an end-to-end manner. We design our network with transformer networks to learn globally and locally varying motions from encoded features of a motion-blurred input, and decode left and right frame features without explicit frame supervision. A flow estimator network is then used to estimate optical flow from the decoded features in a coarse-to-fine manner. We qualitatively and quantitatively evaluate our model through a large set of experiments on synthetic and real motion-blur datasets. We also provide in-depth analysis of our model in connection with related approaches to highlight the effectiveness and favorability of our approach. Furthermore, we showcase the applicability of the flow estimated by our method on deblurring and moving object segmentation tasks.

\end{abstract}

\section*{Introduction}
Acquiring an image with a camera requires the photosensitive sensor of the camera to be exposed to light for a certain duration in order to collect enough photons. Therefore, if a rapid motion of the camera is performed during this time interval -- or if a dynamic object is present in the scene, the resulting image will be \emph{motion-blurred} and will appear smeared along the direction of the relative motion. Motion blur is often considered to be a degrading artifact in many computer vision applications , such as, 3D reconstruction~\cite{seok2013dense} and visual SLAM~\cite{lee2011simultaneous}. Therefore, existing deblurring approaches have been proposed to restore a clean image from a blurred image \cite{zheng2013forward,hyun2014segmentation,wieschollek2016end,gong2017blur2mf}.
 
On the other hand, instead of being considered as an unwanted noise that is to be removed, it has been shown that motion-blur in an image may have some practical interest in core computer vision problems, such as optical flow estimation~\cite{rekleitis1995visual,schoueri2009optical,dai2008motion}, video sequence restoration~\cite{Jin_2018_CVPR,purohit2019bringing}, and 3D scene reconstruction~\cite{Qiu_2019_CVPR}.

Our work focuses on recovering the apparent motion of camera and objects during the exposure time from a single motion-blurred input \ie optical flow estimation. Earlier works~\cite{rekleitis1995visual,schoueri2009optical} deployed a uniform motion-blur assumption, which is often violated in practice. To address this issue, follow-up works~\cite{portz2012optical,li2014robust,tu2015estimating,li2014robust} extended the traditional warping-based methods to a sequence of blurred inputs by imposing further constraints such as non-uniform blur matching and blur gradient constancy. Compared to earlier works, we estimate dense optical flow only from a single motion-blurred image in an end-to-end manner using a novel deep learning framework. Moreover, our method does not impose any restrictive assumption on the motion-blur, hence, it is robust for various blur types.

Another line of work closely related to our work is motion-flow estimation from a blurred image for deblurring applications \cite{dai2008motion,hyun2014segmentation,gong2017blur2mf}. However, these works assume restrictive assumptions for motion estimation, \eg, linear motion assumption~\cite{dai2008motion,hyun2014segmentation} and constrained flow magnitude and direction~\cite{gong2017blur2mf}.
All related works so far imposed a constraint on the motion-kernel, and hence experimented with synthetically simulated blurs under the imposed constraint. In comparison, we train and analyze our network with blurs generated from real high speed videos with no particular motion assumption.

Our proposed framework is composed of three network components: feature encoder, feature decoder, and flow estimator. The feature encoder extracts features from the given motion-blurred input at different spatial scales in a top-down fashion. A feature decoder then decodes the extracted features in a bottom-up manner by learning motion from blur. The feature decoder is composed of spatial transformer networks (STNs)~\cite{jaderberg2015spatial} and feature refining blocks to learn globally and locally-varying motions from the encoded features, respectively. The flow estimator inputs the decoded features at different levels and estimates optical flow in a coarse-to-fine manner. 

We conduct experiments on synthetic and real image blur datasets.
The synthetic image motion-blur dataset is generated via frame interpolation~\cite{slomo} using the Monkaa dataset~\cite{MIFDB16}. For real image motion-blur datasets, we follow previous studies \cite{Nah_2017_CVPR,Jin_2018_CVPR} and temporally average sequential frames in high speed videos. 
As ground truth optical flow does not exist for the high speed video datasets, we compute a flow between the first and last end frames (from which the motion blurred image is averaged) using pretrained state-of-the-art optical flow models \cite{Sun2018PWC-Net,IMKDB17} and use the computed flow as a pseudo-supervision (pseudo-ground truth).

Our contributions are summarized as follows: (1). To the best of our knowledge, we present the first deep learning solution to estimate dense optical flow from a motion-blurred image without any restrictive assumption on the motion kernel. (2). We use pseudo-ground truth flow obtained from pretrained models to successfully train our network for the optical flow estimation from a motion-blurred image. We also show the effectiveness of the proposed network in inferring optical flow from synthetic and real image motion-blur datasets through in-depth experiments. 
(3). We perform detailed analysis of our model in comparison to related approaches and perform ablation studies on different network components to show the effectiveness and flexibility of our approach. (4). We showcase the applicability of the optical flow estimated by our method on motion-blur removal and moving object segmentation tasks.

\section{Related works}

\paragraph{Optical flow and motion-blur.}
Brightness constancy and spatial smoothness assumptions \cite{horn1981determining} often do not conform when estimating optical flow in a blurry scenario. To cope with this limitation, Rekleitis \etal~\cite{rekleitis1995visual} proposed an algorithm to estimate optical flow from a single motion-blurred gray-scale image based on an observation that motion-blur introduces a ripple in the Fourier Transform of the image.
Schoueri \etal~\cite{schoueri2009optical} extended this algorithm to colored images by computing the weighted combination of the optical flow of each image channel estimated by~\cite{rekleitis1995visual}. These methods~\cite{rekleitis1995visual,schoueri2009optical}, however, were limited to linear deblurring filters. To deal with spatially-varying blurs, multiple image sequences were exploited by adapting classical warping-based approaches with modified intensity \cite{portz2012optical} and blur gradient \cite{li2014robust} constancy terms. However, these assumptions are often limited when extended to real motion-blurred images. In this work, we introduce a deep learning framework to estimate  optical flow from a single motion-blurred image without any particular motion assumption, and our proposed model generalizes well for real motion-blurred examples.

\paragraph{Motion flow for deblurring.}
Some of previous deblurring approaches also estimate the underlying motion in a motion-blurred image. For example,~\cite{cho2009fast,dai2008motion,fergus2006removing,zheng2013forward} assume uniform motion blur to remove a blur from an image; however, these methods often fail to remove non-uniform motion blurs. To address this issue, non-uniform deblurring approaches~\cite{gupta2010single,wieschollek2016end,levin2007blind,pan2016soft,hyun2014segmentation} used motion priors based on specific motion models. Since real motion-blurred images often do not comply to these motion assumptions, learning-based discriminative approaches~\cite{chakrabarti2010analyzing,couzinie2013learning} have been proposed to learn motion patterns from a blurred image. These methods are, however, limited since features are manually designed with simple mapping functions. Recently, Gong \etal~\cite{gong2017blur2mf} proposed a deep learning approach for heterogeneous blur removal via motion-flow estimation. They treated motion-flow prediction as a classification task and trained a fully convolutional network over a discrete output domain.

\paragraph{Recent works on learning from motion-blur.}
 Our work is also related to recent deep learning approaches that focus on unfolding the hidden information in a motion blurred image. Jin \etal~\cite{Jin_2018_CVPR} and Purohit \etal~\cite{purohit2019bringing} reconstructed latent video frames from a single motion-blurred image. Qui \etal~\cite{Qiu_2019_CVPR} proposed a method to recover the 3D scene collapsed during the exposure process from a motion-blurred image. These works show that motion blur can become meaningful information when processed properly. In this paper, in addition to the previously proposed applications, we explore another potential of motion blur, \ie optical flow estimation.
\begin{figure*}[!t]
\begin{center}
\setlength{\tabcolsep}{0.4pt}
\resizebox{1.0\linewidth}{!}{
\begin{tabular}{ccccc}
         \includegraphics[width=0.1\linewidth]{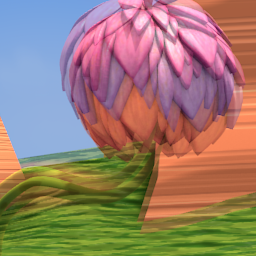} &
         \includegraphics[width=0.1\linewidth]{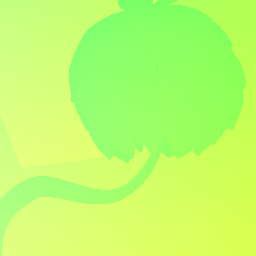} &
         \includegraphics[width=0.1\linewidth]{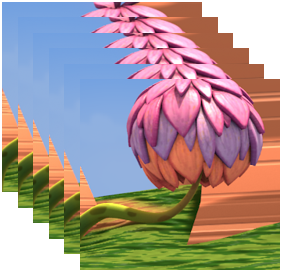} &
         \includegraphics[width=0.1\linewidth]{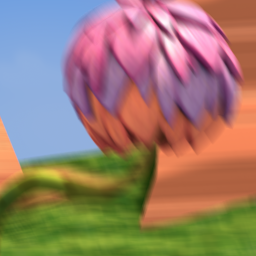} &
         \includegraphics[width=0.1\linewidth]{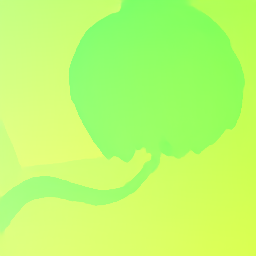}\\
         \scriptsize(a) & \scriptsize(b) & \scriptsize(c) & \scriptsize(d) & \scriptsize(e) 
         \end{tabular}}
\end{center}
\vspace{-2mm}
\caption{(a) First and last images from \cite{MIFDB16} overlaid (b) Ground truth optical flow between first and last images (c) Intermediate frame interpolation using \cite{slomo} (d) Motion-blurred image generated by averaging first, intermediate and last images (e) Optical flow predicted by our model from the motion-blurred image.}
\label{fig:data_generation}
\vspace{-2mm}
\end{figure*}

\section{Dataset generation}
\label{sec:data_generation}
Manually collecting a large set of blurred images is challenging and requires a great deal of human effort. Hence, a common practice in computer vision research is to generate motion-blurred images by averaging sequential frames in high frame rate videos~\cite{Jin_2018_CVPR,purohit2019bringing,Nah_2017_CVPR}. 
 In this work, we experiment with both synthetic and real scene blur datasets without particular assumption, and the process of generating these datasets is described below.

\paragraph{Synthetic scene blur dataset.}
We take advantage of the Monkaa dataset proposed in~\cite{MIFDB16} to generate a synthetic image motion-blur dataset for optical flow estimation. The dataset provides ground truth flows between a pair of frames in synthetic video scenes. Given two consecutive frames, in order to simulate the motion blur, we interpolate intermediate frames using~\cite{slomo} and average all resulting frames. 
The number of intermediate frames to be interpolated relies on the magnitude of the motion of objects in the scene to ensure a smooth motion-blur generation. Hence, to generate a natural-like motion blur, we defined a simple relationship between the number of frames to be interpolated $n$ and the maximum pixel displacement $s$ in the ground truth flow as follows: $n = \mathrm{max}(|s|,5)$, where $|s|$ is the absolute value of the $s$. To avoid severely blurred images, we discarded samples with $|s| > 100$. We generate a Monkaa blur dataset with 10,000 training and 1200 test images (see \Fref{fig:data_generation}).

\paragraph{Real scene blur dataset.}
To generate real scene motion-blur images for network training, we use high speed video datasets: GoPro \cite{Nah_2017_CVPR} and NfS \cite{galoogahi2017need}. The GoPro high speed video dataset which is commonly used for dynamic scene deblurring has 33 videos taken at 240fps, out of which 25 videos are used for training and the rest are used for testing. Motion-blurred images are generated by averaging 7 consecutive frames in a video. As a result, we obtain blurred frames approximately 30 fps which is common setting for commercial cameras. We refer to the blur dataset generated from~\cite{Nah_2017_CVPR} as the GoPro blur dataset. We also experimented with the Need for Speed (NfS)~\cite{galoogahi2017need} high speed video dataset, a common benchmark for visual tracking task. The dataset contains more diverse scenes for large-scale training. It is also a better fit to the task of estimating flow as most of the videos contain dynamic motion of objects in a scene with a close to static background compared to~\cite{Nah_2017_CVPR} where egomotion is predominant. Out of 100 videos in the dataset, 70 are used for training and the remaining videos are used for validation and testing. We call the motion-blur dataset generated from~\cite{galoogahi2017need} as the NfS blur dataset. In case of real blur datasets, the optical flow between the first ($1^{\textrm{st}}$) and last ($7^{\textrm{th}}$) frames, which is used as a self-supervision during training, is obtained using pretrained state-of-the-art models~\cite{Sun2018PWC-Net,IMKDB17}.


\section{Methodology}
\begin{figure*}[!t]
	\centering
	\includegraphics[width=1\linewidth,trim={8.5cm 2cm 7cm 4.5cm},clip]{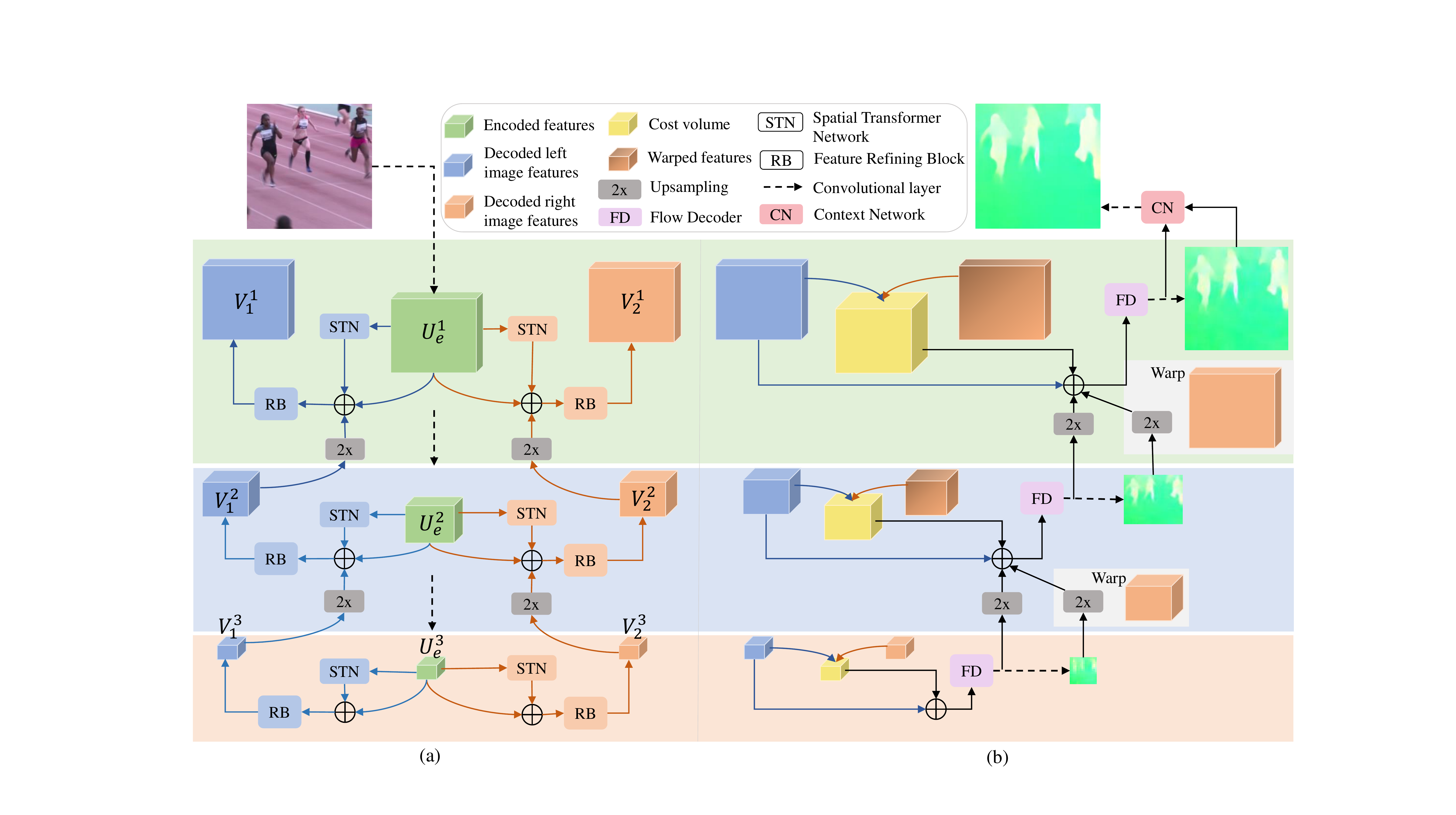}
\caption{Overview of our network. (a) Feature encoding and decoding  (b) Flow estimation. Our network estimates flows at 6 different scales. We only visualize 3 levels for simplicity  (Best viewed in color).}
	\label{fig:model}

\end{figure*}

In this section, we explain our framework and details of the training process. Our model has 3 main components: A feature encoder, feature decoder and flow estimator (see \Fref{fig:model}).

\paragraph{Feature encoder.}
The feature encoder extracts features at different scales from a given motion-blurred image. It is a feed-forward convolutional network which has six convolutional blocks each with two layers of convolutions of kernel size $3\times3$ and stride size of 2 and 1, respectively, with a ReLU nonlinearity following each convolutional layer. During the feature extraction stage, features are downsampled to half of their spatial size after each convolutional block. Given an input image $I$, the feature encoder outputs $k$ features: $\{U_e^l\}_{l=1}^{k}$, where $U_e^l$ is an encoded feature at level $l$ (see \Fref{fig:model}a).

\paragraph{Feature decoder.}
The feature decoder is composed of spatial transformer networks (STNs)~\cite{jaderberg2015spatial} and feature refining blocks to decode the encoded features into first and last frame features in a bottom-up fashion. Given a feature $U_e^l \in \mathbb{R} ^{H\times W\times C}$ with width $W$, height $H$ and $C$ channels from the encoder, the STN predicts a global transformation parameters $\theta$ conditioned on the input to spatially transform the feature, hence, learning non-local motion from the feature of a motion-blurred input (\Eref{eqn:stn}). 
In order to take account of locally variant motions which are apparent in dynamic scenes, the transformed feature is passed through a refining network. Along with the transformed feature, we input the feature from the encoder into the refining network in order to guide the network to infer the relative spatial motion (see \Fref{fig:model}a). A residual connection is also built by upsampling decoded features from previous feature level using a transposed convolution (deconvolution) layer of kernel size $4\times4$ and stride size of 2.

At each feature level $l$, the refining network $\calR^l$ takes the transformed feature $U_t^l$, the encoded feature $U_e^l$, and the upscaled decoded feature from the previous feature level $l+1$ concatenated together channel-wise and outputs a decoded feature (\Eref{eqn:feat_dec}). The refining network contains five densely connected convolutional layers each with kernel size $3\times3$ and stride size 1. We used individual feature decoders to reconstruct first and last image features at different levels of abstraction. These features are then used to estimate optical flow, mimicking the vanilla pipeline for optical flow estimation from two images as shown in \Fref{fig:model}b.

\begin{equation}
    U_t^l = \calT^l_\theta\{U_e^l\}
    \label{eqn:stn}
\end{equation}
\begin{equation}
    V^l = \calR^l\{U_t^l \oplus U_e^l \oplus \mathbf{up.}(V^{l+1})\}
    \label{eqn:feat_dec}
\end{equation},
where $l=\{1,\; ...\; ,k\}$, $\calT_\theta$ denotes transformation by STN, $\mathbf{up.}$ denotes upsampling, $\oplus$ denotes channel-wise concatenation, and $V^l$ is the decoded feature at level $l$.

\paragraph{Flow estimator.}
The flow estimator computes optical flow at different scales using the decoded first and last frame features. Note that the decoded features here are equivalent to the encoded features of two clean input images in standard optical flow estimation algorithms. Inspired by recent network architectures for optical flow estimation~\cite{fischer2015flownet,IMKDB17,ranjan2017optical,Sun2018PWC-Net}, we reconstruct flow maps from coarse-to-fine using cost volume, warping, flow decoding, and prediction layers. At each feature level $l$, the warping layer $\calW$ warps the last frame feature $V_2^l$ at level $l$ with an upsampled flow estimated at previous scale $l+1$ (\Eref{eqn:warp}). A deconvolution layer of kernel size $4\times4$ and stride size of 2 was used to upsample flows at different scales. 

Given a first image feature $V_1^l$ and a backward-warped last image feature $\hat{V}_2^l$, the cost volume layer $\calC$ computes the matching cost between the features using a correlation layer~\cite{fischer2015flownet,xu2017accurate,Sun2018PWC-Net}. The flow decoding network $\calD$ takes the output of the cost volume layer and decodes a flow feature $V_f^l$ that will be used by the flow prediction layer to estimate flow.  In addition to the correlation output, the flow decoder takes the first image feature, the upscaled flow and decoded flow feature from previous feature level (\Eref{eqn:flow_dec}). Like the refining block in the image feature decoder, the flow decoder has five densely connected convolutional layers each with kernel size $3\times3$ and stride size 1. 

The flow prediction layer $\calP$ inputs the decoded flow feature and predicts optical flow. It is a single convolutional layer that outputs a flow $f^l$ of size $H \times W\times2$ given a feature $V_f^l \in \mathbb{R} ^{H\times W\times C}$. Flow maps are estimated at different scales from the smallest to the highest resolution (\Eref{eqn:flow_pred}). The estimated full-scale flow is further refined by aggregating contextual information using a context network~\cite{Sun2018PWC-Net,im2018dpsnet}. 
It contains seven dilated convolutions~\cite{yu2015multi} with receptive field size of 1,2,4,8,16,1 and 1, respectively. The context network takes the final decoded flow feature and the predicted flow as an input and outputs a refined flow. Each convolutional layer in the flow decoder and flow estimator is followed by a ReLU activation layer. 

\begin{equation}
    \hat{V}_2^l = \calW\{V_2^l, \mathbf{up.}(f^{l+1})\}
    \label{eqn:warp}
\end{equation}
\begin{equation}
    V_f^l = \calD^l\{\calC(V_1^l,\hat{V}_2^l) \oplus V_1^l  \oplus \mathbf{up.}(V_f^{l+1}))\oplus \mathbf{up.}(f^{l+1})\}
    \label{eqn:flow_dec}
\end{equation}
\begin{equation}
    f^l = \calP^l\{V_f^l\}
    \label{eqn:flow_pred}
\end{equation}

\begin{figure*}[!t]
\begin{center}
\setlength{\tabcolsep}{0.8pt}
\renewcommand{\arraystretch}{0.6}
\resizebox{1.0\linewidth}{!}{
    \begin{tabular}{cccccc}
         \scriptsize{Input} & \scriptsize{GT} & \scriptsize{Ours} & \scriptsize{Input} & \scriptsize{GT} & \scriptsize{Ours} \\
         \includegraphics[width=0.1\linewidth]{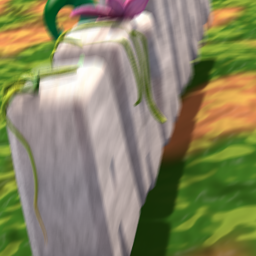} &
         \includegraphics[width=0.1\linewidth]{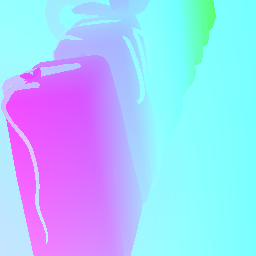} &
         \includegraphics[width=0.1\linewidth]{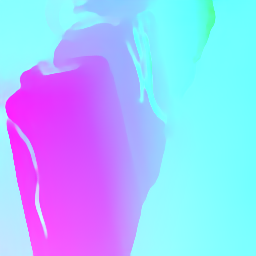} &
         \includegraphics[width=0.1\linewidth]{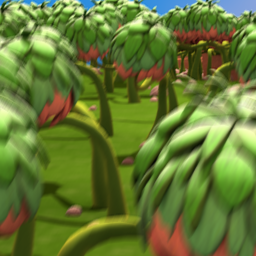} &
         \includegraphics[width=0.1\linewidth]{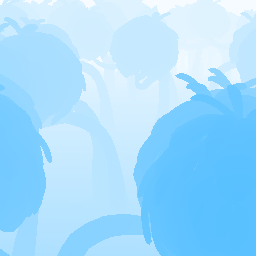} &
         \includegraphics[width=0.1\linewidth]{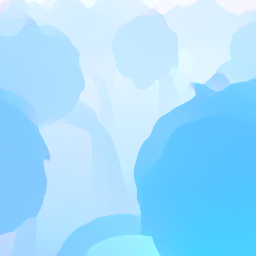}
    \end{tabular}}
\end{center}
\caption{Qualitative results on Monkaa blur dataset. The $1^\mathrm{st}$ and $4^\mathrm{th}$ columns are motion-blurred inputs generated by temporally averaging intermediate frames interpolated between first and last images in \cite{MIFDB16}. The $2^\mathrm{nd}$ and $5^\mathrm{th}$ columns are ground truth flows. The $3^\mathrm{rd}$ and $6^\mathrm{th}$ columns are optical flows estimated by our model given the motion-blurred input.}
\label{fig:monkaa_result}
\end{figure*}

\begin{figure*}[!t]
\begin{center}
\setlength{\tabcolsep}{0.8pt}
\renewcommand{\arraystretch}{0.6}
\resizebox{1\linewidth}{!}{
\begin{tabular}{cccccc}
         \scriptsize{Input} & \scriptsize{p-GT} & \scriptsize{Ours} & \scriptsize{Input} & \scriptsize{p-GT} & \scriptsize{Ours} \\
         \includegraphics[width=0.1\linewidth]{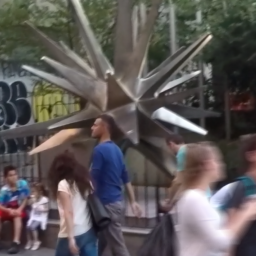} &
         \includegraphics[width=0.1\linewidth]{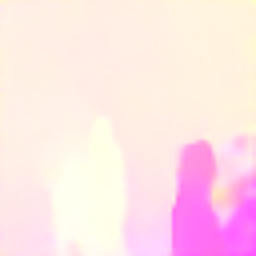} &
         \includegraphics[width=0.1\linewidth]{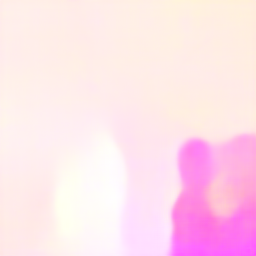}&
         \includegraphics[width=0.1\linewidth]{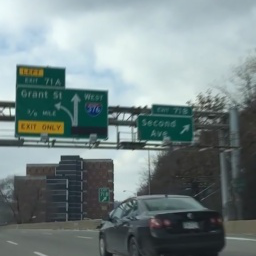} &
         \includegraphics[width=0.1\linewidth]{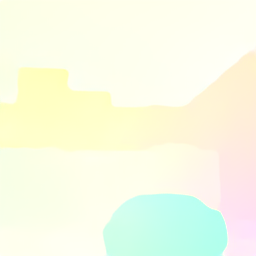} &
         \includegraphics[width=0.1\linewidth]{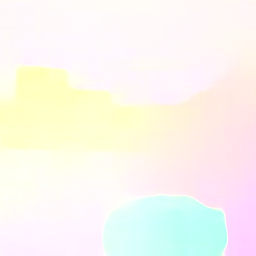} \\
         \includegraphics[width=0.1\linewidth]{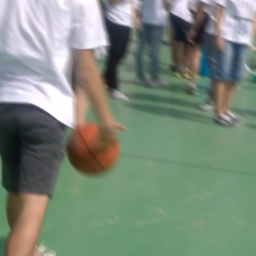} &
         \includegraphics[width=0.1\linewidth]{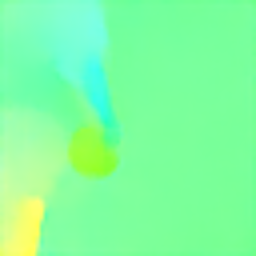} &
         \includegraphics[width=0.1\linewidth]{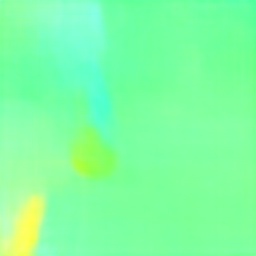} &
         \includegraphics[width=0.1\linewidth]{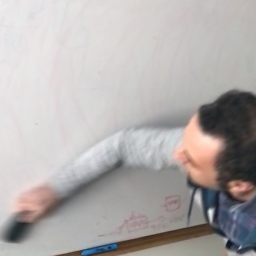} &
         \includegraphics[width=0.1\linewidth]{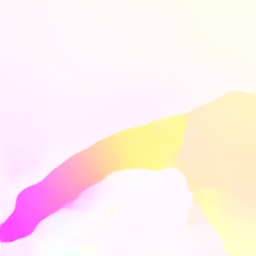} &
         \includegraphics[width=0.1\linewidth]{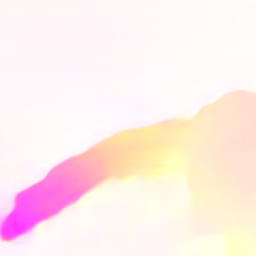} 
\end{tabular}}
\end{center}
\caption{Qualitative results on GoPro and NfS blur datasets. We compare the optical flow estimated by our model with pseudo-ground truth optical flow (p-GT) between the first and last frames in a high-speed video sequence (predicted using pretrained optical flow network \cite{Sun2018PWC-Net,IMKDB17}) from which the motion blurred image was generated.}
\label{fig:gopro_result}
\vspace{-3mm}
\end{figure*}

\paragraph{Network training.}
Extracting motion information from a single motion-blurred image is an ill-posed problem without extra information (from external sensors such as IMU or other clues on the camera motion \cite{lee2019visuomotor}) as averaging destroys temporal order~\cite{Jin_2018_CVPR,purohit2019bringing,Qiu_2019_CVPR}.  We experimentally find that a weighted multi-scale endpoint error (EPE) between the estimated flows and the downscaled pseudo-ground truth flows is a sufficient constraint for network training and convergence (\Eref{eqn:loss}). We used a bilinear interpolation to downsample the pseudo-ground truth flow to respective sizes at different scales. An attempt to use $\ell1$ photometric loss for supervision or smoothness loss as a regularizer did not improve network performance. 

\begin{equation}
    \calL = \sum_{l=1}^{k}w_l \cdot\big|f_l -\hat{f}_{l}\big|_2
    \label{eqn:loss}
\end{equation}

, where $f_l$ is the estimated flow, $\hat{f}_l$ is the downsampled  pseudo-ground truth flow and $w_l$ is the loss weight coefficient at scale $l$.

Given a pseudo-ground truth optical flow as a supervision during training, our network learns to decode the first and last image features from encoded features of an input image in a symmetric manner. The decoded features are then used to estimate flow. This task, however, can potentially suffer from an ambiguity in predicting the correct flow direction during inference. Given a single motion-blurred image, estimating the flow direction (either from first$\rightarrow$ last or last $\rightarrow$ first) is a highly intractable problem as reversely averaging sequential frames result in the same motion-blurred image. Hence, to purely measure the quality the predicted flows without flow direction estimation issues, we evaluate both forward and backward direction flows and report the lower EPE in the experiment section. Despite the existence of temporal ambiguity, optical flow from a motion-blurred image has many practical applications in computer vision research such as 3D reconstruction, segmentation from flow and motion-blur removal (see Downstream tasks).

\section{Experiment}
\begin{table}[!t]
\setlength{\tabcolsep}{6pt}
\begin{center}
\caption{Quantitative evaluation. For simplicity, we refer to our Blur to Flow network as B2F-net.}

\label{tbl:quant_result}
\begin{tabular}{l|ccc}
\toprule
Method & Monkaa & GoPro & NfS \\
\midrule
B2F-net & 1.158 & 2.077 & - \\
B2F-net + \emph{fine tuning} & - & 2.038 & 1.958\\
\bottomrule
\end{tabular}
\end{center}
\vspace{-3mm}
\end{table}
\paragraph{Implementation details.}
We estimate flows at 6 different feature levels with training loss weight coefficients set as follows: $w_6 = 0.32$, $w_5 = 0.08$, $w_4 = 0.04$, $w_3 = 0.02$, $w_2 = 0.01$ and $w_1 = 0.005$ from the lowest to the highest resolution, respectively. At each level, we use a correlation layer with a neighborhood search range of 4 pixels and stride size of 1. We chose Adam~\cite{KingmaB14} as an optimization method with parameters $\beta_1$, $\beta_2$ and \emph{weight decay} fixed to 0.9, 0.999 and $4e-4$, respectively. In all experiments, a mini-batch size of 4 and image size of $256\times256$ is used by centrally cropping inputs. Following~\cite{fischer2015flownet}, we train the Monkaa blur dataset for 300 epochs with initial learning rate $\lambda = 1e-4$. We gradually decayed the learning rate by half at 100, 150, 200 and 250 epochs during training. For the GoPro and NfS blur datasets, we trained (fine-tuned) the model for 120 epochs with a learning rate initialized with $\lambda = 1e-4$ and decayed by half at 60, 80 and 100 epochs. 

\paragraph{Qualitative evaluation.}
For the Monkaa~\cite{MIFDB16} blur dataset, we compare our results with the ground truth optical flow between the first and last frames which we used to interpolate intermediate frames and synthesize a motion-blurred image. Our model successfully estimates optical flow from synthetic blurred inputs with different blur magnitudes and patterns (see \Fref{fig:monkaa_result}). \Fref{fig:gopro_result} shows test results on the real image blur datasets~\cite{Nah_2017_CVPR,galoogahi2017need}. In order to evaluate whether our model reasonably inferred the flow from a given real motion-blurred image, our results are compared with the optical flow between sharp initial and final frames in a high speed video (from which the blurred image was temporally averaged) predicted by pretrained state-of-the-art models~\cite{Sun2018PWC-Net,IMKDB17}. These flows are later used as pseudo-ground truth (p-GT) for quantitative evaluation. The qualitative results on real image blur datasets show that our model accurately predicts the motion of objects in different motion-blur scenarios with dynamic motion of multiple objects in a close to static or moving scenes. Failure cases occur for temporally undersampled and heavily blurred samples since image contents of such samples are often destroyed and, hence, feature decoding usually fails. 

\paragraph{Quantitative evaluation.}
We compute the end-point error between the estimated optical flow from a motion-blurred input and the flow between the sharp first and last images (pseudo-ground truth). Due to flow direction ambiguity, we calculate both the forward and backward flows during testing and report the lower metric \ie $\mathrm{min}\{\mathrm{EPE}(f_{1\rightarrow2}), \mathrm{EPE}(f_{2\rightarrow1})\}$. The averaged test results on different datasets are summarized in \Tref{tbl:quant_result}. The test error for the Monkaa blur dataset is lower compared to real datasets as the simulated blurs from the synthetic scenes are mostly static. We experimentally find that training our model from scratch with random weight initialization converges well on the GoPro blur dataset, but  does not converge on the NfS blur dataset. This is mainly because the NFS blur dataset contains various low quality videos with some of them containing out-of-focus frame sequences, and hence the resulting blur dataset is challenging for our network to learn. To mitigate this issue, we used a model pretrained on the synthetic dataset and fine-tuned the model on the NfS blur dataset. Training in this manner resulted in a good network performance. Moreover, fine-tuning also lowered the endpoint error on the GoPro blur dataset (see \Tref{tbl:quant_result}). 

\begin{figure}[!t]
\begin{center}
\setlength{\tabcolsep}{0.4pt}
\renewcommand{\arraystretch}{0.3}
\resizebox{1\linewidth}{!}{%
\footnotesize
\begin{tabular}{ccccc}
        \includegraphics[width=0.1\linewidth]{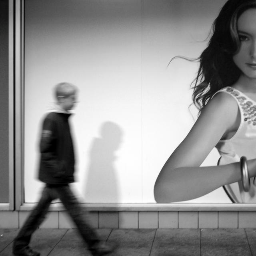} &
        \includegraphics[width=0.1\linewidth]{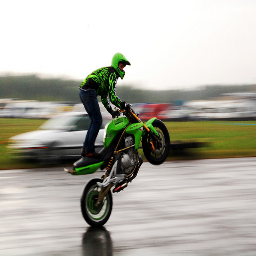} &
        \includegraphics[width=0.1\linewidth]{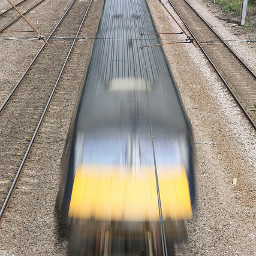} &
        \includegraphics[width=0.1\linewidth]{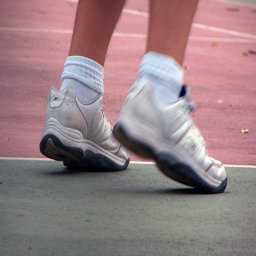} \\
        \includegraphics[width=0.1\linewidth]{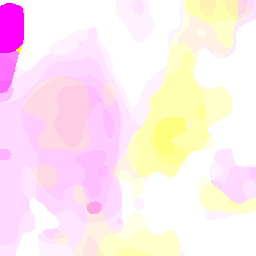} &
        \includegraphics[width=0.1\linewidth]{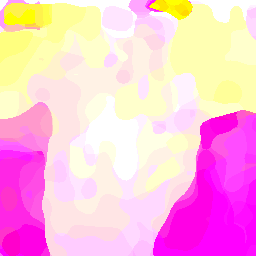} &
        \includegraphics[width=0.1\linewidth]{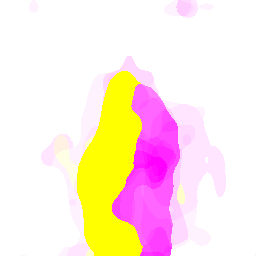} &
        \includegraphics[width=0.1\linewidth]{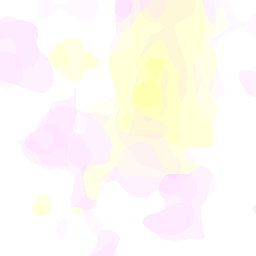}\\  
        \includegraphics[width=0.1\linewidth]{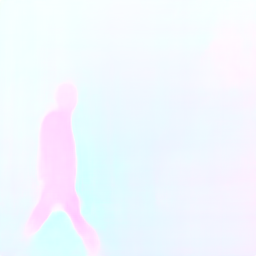} &
        \includegraphics[width=0.1\linewidth]{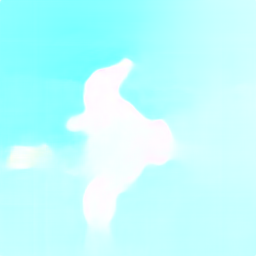} &
        \includegraphics[width=0.1\linewidth]{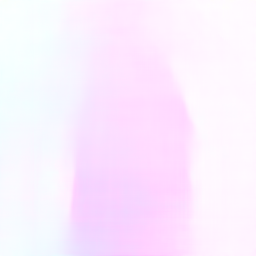} &
        \includegraphics[width=0.1\linewidth]{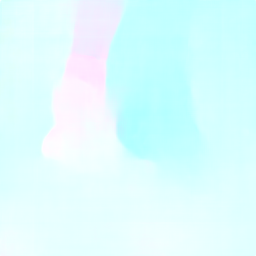}  
\end{tabular}}
\end{center}
\caption{Comparison with motion flow estimation works. The $2^\mathrm{nd}$ and $3^\mathrm{rd}$ rows depict  outputs of \citeauthor{gong2017blur2mf} and our model, respectively.}
\label{fig:qual_motion_flow}
\vspace{-3mm}
\end{figure}

\section{Analysis}
\paragraph{Motion flow estimation.}
We qualitatively compare our method with previous works that estimate motion flow from a blurred image for deblurring purpose \cite{hyun2014segmentation,gong2017blur2mf}. In order to perform a fair comparison, both our model and the model from \cite{gong2017blur2mf} are evaluated on motion-blurred examples from a real blur dataset~\cite{jianping2014}. As can be inferred from \Fref{fig:qual_motion_flow}, our model generalizes noticeably well by estimating a reasonable flow from the given blurred inputs. On the other hand, it is hardly possible to analyse the predictions from~\cite{gong2017blur2mf} as their model fails to output an interpretable optical flow. This is mainly because~\cite{gong2017blur2mf} treats the task of estimating flow from a motion-blurred image as a classification problem and predicts discrete integer vectors at each pixel with constrained flow direction and magnitude. Moreover, due to the nature of the problem setup, their network could be trained only on synthetic images with simulated motion blurs, hence making it difficult to generalize for real motion-blurred cases. In regard to these two aspects, our model treats optical flow estimation from a motion-blurred image as a regression problem with no specific constraint and can be trained on motion-blurred images from real high speed videos, which leads to the better generalization capability of our model.

\begin{figure}[!t]
\begin{center}
Input\hspace{1.3cm}p-GT\hspace{0.95cm}Jin $\rightarrow$ flow \hspace{0.8cm}Ours
\setlength{\tabcolsep}{0.4pt}
\renewcommand{\arraystretch}{0.3}
\resizebox{1.0\linewidth}{!}{%
\begin{tabular}{cccc}
        \includegraphics[width=0.1\linewidth]{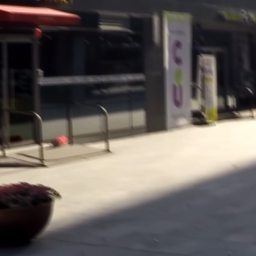} &
        \includegraphics[width=0.1\linewidth]{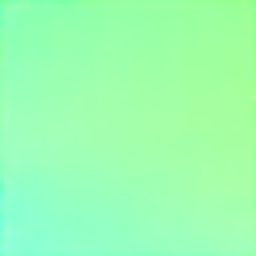} &
        \includegraphics[width=0.1\linewidth]{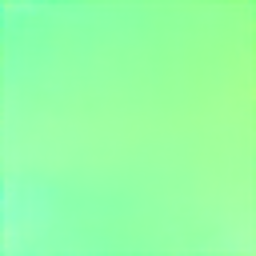} &
        \includegraphics[width=0.1\linewidth]{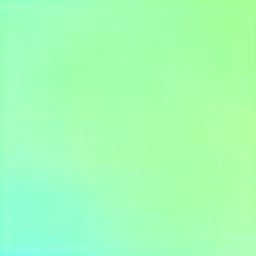} \\
        \includegraphics[width=0.1\linewidth]{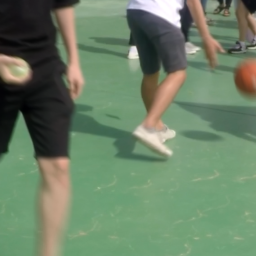} &
        \includegraphics[width=0.1\linewidth]{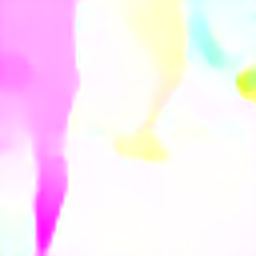} &
        \includegraphics[width=0.1\linewidth]{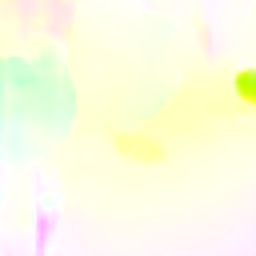} & 
        \includegraphics[width=0.1\linewidth]{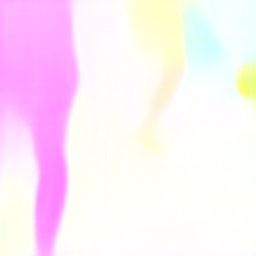} \\
        \includegraphics[width=0.1\linewidth]{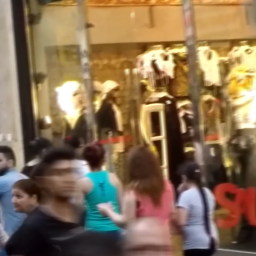} & 
        \includegraphics[width=0.1\linewidth]{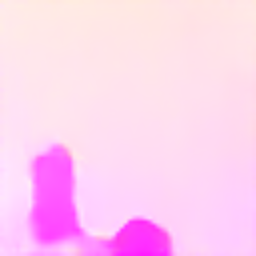} & 
        \includegraphics[width=0.1\linewidth]{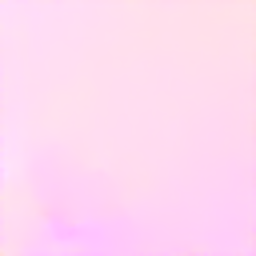} & 
        \includegraphics[width=0.1\linewidth]{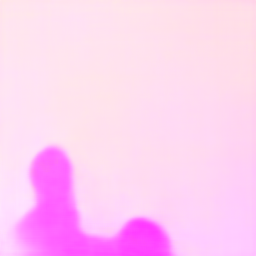} 
\end{tabular}}
\end{center}

\caption{Comparison with flow from restored frames. The $3^\mathrm{rd}$ column depicts the flows computed between the first and last frames in the restored sequence \cite{Jin_2018_CVPR}. The $4^\mathrm{th}$ column shows our model prediction.}
\label{fig:qual_cvpr18}
\vspace{-3mm}
\end{figure}

\paragraph{Flow from restored frames.}
Following recent works in video sequence reconstruction from a motion-blurred image~\cite{Jin_2018_CVPR,purohit2019bringing}, a naive approach to the task at hand would be to restore left and right end frames and estimate optical flow from the restored frames using standard approaches. Here we compare the optical flow of a motion-blurred input estimated by our model with a flow predicted by PWC-Net  from restored left and right frames using~\cite{Jin_2018_CVPR}. Quantitatively, this approach performs considerably worse giving an endpoint error of magnitude 5.875 on the GoPro blur dataset compared to our approach (2.077). The qualitative results in \Fref{fig:qual_cvpr18} also show that our approach outputs more accurate results. This performance gap can be directly attributed to the fact that restored frames usually contain multiple motion artifacts. Moreover, we experimentally find that estimating flow from restored frames works relatively well for uniform motion blurs ($1^\mathrm{{st}}$ row in \Fref{fig:qual_cvpr18}) since sequence restoration methods like Jin \etal generally learn the global camera motion in a static scene. However, for dynamic blurs ($2^\mathrm{{nd}}$ and $3^\mathrm{{rd}}$ rows in \Fref{fig:qual_cvpr18}), these methods likely output inaccurate optical flows as they often fail to correctly capture locally varying motions. 
On the contrary, our method successfully captures global and local motions.

\section{Downstream tasks}
\begin{figure}[!t]
\begin{center}
{\hspace{0.3cm}Input\hspace{0.8cm}Appearance\hspace{0.8cm}Motion\hspace{1.2cm}Joint}

\setlength{\tabcolsep}{0.4pt}
\renewcommand{\arraystretch}{0.3}
\resizebox{1.0\linewidth}{!}{%

\begin{tabular}{cccc}
        \includegraphics[width=0.1\linewidth]{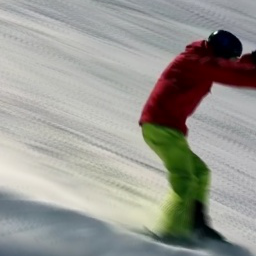} &
        \includegraphics[width=0.1\linewidth]{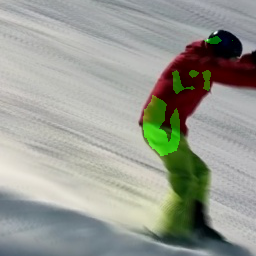} &
        \includegraphics[width=0.1\linewidth]{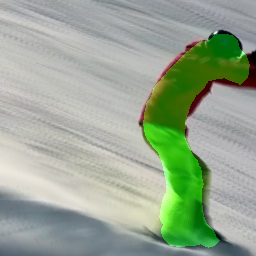} &
        \includegraphics[width=0.1\linewidth]{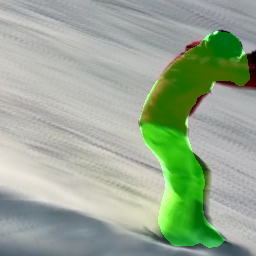} \\
        \includegraphics[width=0.1\linewidth]{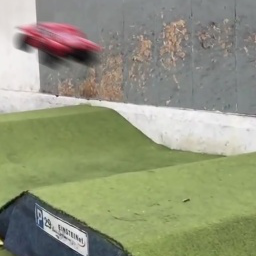} &
        \includegraphics[width=0.1\linewidth]{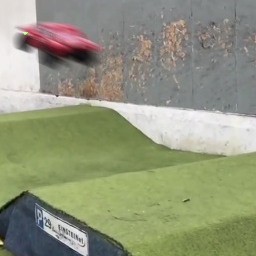} &
        \includegraphics[width=0.1\linewidth]{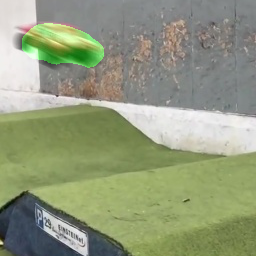} &
        \includegraphics[width=0.1\linewidth]{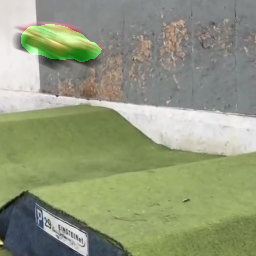} \\
        \includegraphics[width=0.1\linewidth]{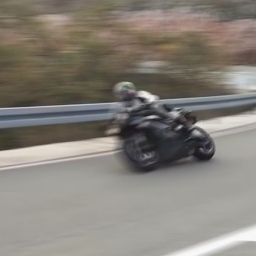} &
        \includegraphics[width=0.1\linewidth]{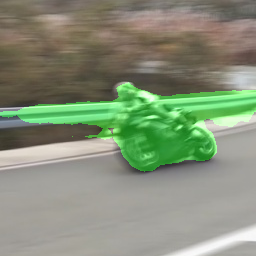} &
        \includegraphics[width=0.1\linewidth]{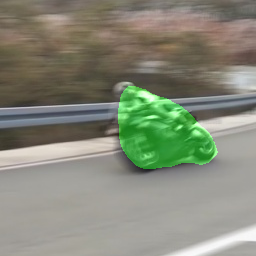} &
        \includegraphics[width=0.1\linewidth]{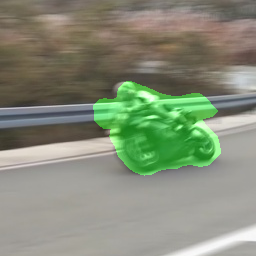} 
\end{tabular}}
\end{center}

\caption{Qualitative analysis on the application of optical flow for moving object segmentation in a blurry scenario.}
\label{fig:segment}
\vspace{-3mm}
\end{figure}

\paragraph{Moving object segmentation.}
Optical flow is commonly used in video object segmentation task along with appearance information to segment moving objects in a video. \cite{jain2017fusionseg,Hu_2018_ECCV,Hu_2018_CVPR}. To highlight the applicability of the flow estimated by our approach, we experimented with the task of segmenting generic moving objects from a motion-blurred input. For this purpose, we used a pretrained FusionSeg model from \cite{jain2017fusionseg} which contains an appearance stream (inputs an image), a motion stream (inputs an optical flow) and a fusion of the two networks. As shown in \Fref{fig:segment}, segmenting moving objects from a motion-blurred input often results in a failure or inaccurate segmentation masks mainly because object boundaries and appearance cues that are crucial for segmentation are corrupted by blur. On the other hand, by feeding the optical flow estimated by our approach into the motion stream network, we obtained accurate segmentation results. The joint model that uses both appearance and motion information also leverages the estimated optical flow to segment the moving object in the given blurred input. This showcases the applicability of the estimated flow for moving object segmentation in a blurry scenario.

\begin{figure*}[!t]
\begin{center}
\setlength{\tabcolsep}{0.8pt}
\renewcommand{\arraystretch}{0.6}
\scriptsize
\resizebox{1.0\linewidth}{!}{%
\begin{tabular}{cccccc}
         \scriptsize{Input (BSD-M)} & \scriptsize{Gong \etal} & \scriptsize{Ours} & \scriptsize{Input (BSD-S)} & \scriptsize{Gong \etal} & \scriptsize{Ours} \\
          \includegraphics[width=0.1\linewidth]{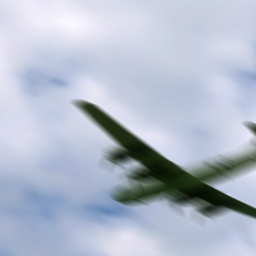}&
          \includegraphics[width=0.1\linewidth]{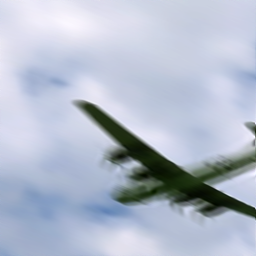}&
          \includegraphics[width=0.1\linewidth]{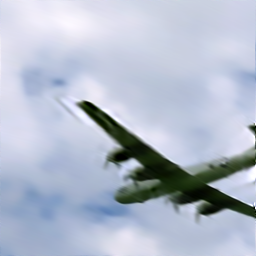}&
          \includegraphics[width=0.1\linewidth]{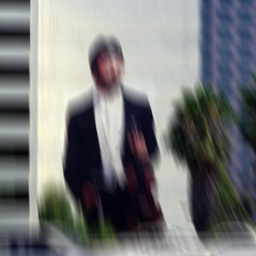}&
          \includegraphics[width=0.1\linewidth]{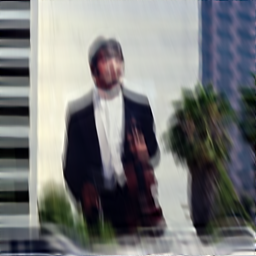}&
          \includegraphics[width=0.1\linewidth]{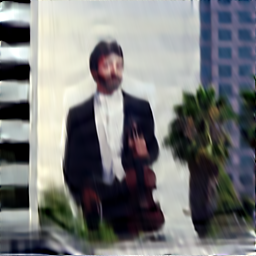}\\
          \includegraphics[width=0.1\linewidth]{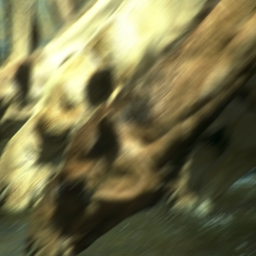}&
          \includegraphics[width=0.1\linewidth]{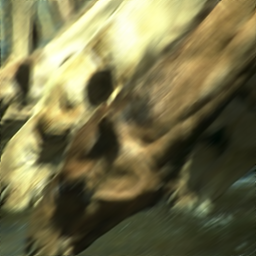}&
          \includegraphics[width=0.1\linewidth]{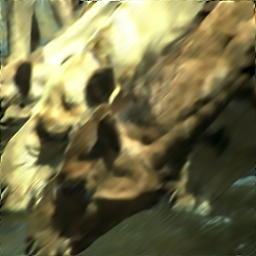}&
          \includegraphics[width=0.1\linewidth]{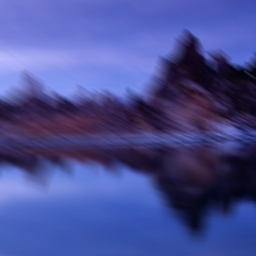}&
          \includegraphics[width=0.1\linewidth]{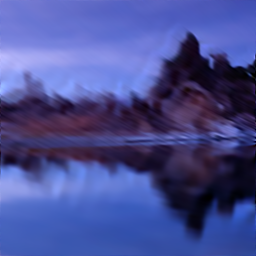}&
          \includegraphics[width=0.1\linewidth]{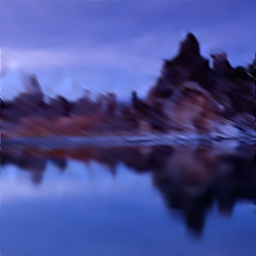}
\end{tabular}}
\end{center}
\caption{Qualitative comparison on motion-blur removal task using the estimated flow.}
\label{fig:supp_blur}
\vspace{-3mm}
\end{figure*}

\paragraph{Motion-blur removal.}
Given a blurred image and the estimated motion kernel, previous works \cite{couzinie2013learning,sun2015learning,gong2017blur2mf} recover a sharp image by performing a non-blind deconvolution algorithm (Please refer to section 3.2 of \citeauthor{gong2017blur2mf} for details). In order to examine the effectiveness of our estimated flows for blur removal, we also directly used our estimated flow in the deblurring process and compared our method with competing approaches \cite{sun2015learning,gong2017blur2mf}. For quantitative comparison, we used the official code of \cite{gong2017blur2mf} to generate two types of motion-blurred test sets: BSD-M (with maximum pixel displacement of 17) and BSD-S (with maximum pixel displacement of 36). Please refer to section 5.1 of \cite{gong2017blur2mf} for dataset details. Each dataset contains 300 motion-blurred images of size $480\times320$. The quantitative and qualitative results are shown in \Tref{tbl:deblur} and \Fref{fig:supp_blur}, respectively. As can be inferred from the results, our approach performs favourably against previous works \cite{sun2015learning,gong2017blur2mf} on recovering sharp frames from a blurred image. These results are a direct consequence of better motion-flow estimation as briefly analyzed in the previous section.

\section{Ablation studies}
\paragraph{Importance of feature decoding.} 
In order to show the importance of the feature decoder in our network, we experimented with a U-net \cite{RFB15a} like architecture where we estimated flow only from encoded features without explicitly decoding left and right image features \ie no spatial transformer networks (STN) and feature refining blocks (RB). A network trained in this manner converged with a higher training error resulting in a worse test error of magnitude 2.748 EPE (32.30\% error increment compared to a model with STN and RB).
\vspace{-1mm}
\paragraph{Spatial transformer network.} 
 STNs~\cite{jaderberg2015spatial} in our network learn non-local motions (\eg camera motion) and estimate global transformation parameters to spatially transform the encoded feature of a given motion-blurred input accordingly. They are one of the building blocks of our network and are notably important for the performance of our model. As can be seen from \Tref{tbl:ablation}, a network with STN lowers the endpoint error by 8.21\% in comparison with a model without STN.
\vspace{-1mm}
\paragraph{Feature refining block.}
The feature refining block (RB) decodes left and right image features for the flow estimator. As the STNs only predict global feature transformation parameters, non-local motions which are predominant in real world scenarios are taken into account by passing the transformed feature in the feature refining block. Our experimental results also verify the significance of feature refining blocks as it can be inferred from \Tref{tbl:ablation}. Training a model without feature refining blocks increments the endpoint error by 14.73\%. 

\begin{table}[!t]
\setlength{\tabcolsep}{4.3pt}
\begin{center}
\caption{Motion-blur removal via non-blind deconvolution}
\label{tbl:deblur}

\begin{tabular}{l|cc|cc}
\toprule
& \multicolumn{2}{c|}{BSD-M} & \multicolumn{2}{c}{BSD-S}  \\ \midrule
& PSNR (dB) & SSIM & PSNR (dB) & SSIM \\ \midrule
Sun \etal & 22.97 & 0.674 & 20.53 & 0.530\\ 
Gong \etal & 23.88&0.718 & 21.85& 0.625\\
Ours & \textbf{25.23} & \textbf{0.786} & \textbf{23.41} & \textbf{0.714}\\ \bottomrule
\end{tabular}
\end{center}
\vspace{-2.5mm}
\end{table}
\begin{table}[!t]
\setlength{\tabcolsep}{15pt}

\begin{center}
\caption{Ablation studies on GoPro blur dataset for different network components}

\label{tbl:ablation}
\begin{tabular}{ccc}
\toprule
STN & RB &  EPE ($\downarrow$)\\
\midrule

\cmark  & \cmark  &  \textbf{2.077} \\
\xmark  & \cmark  &  2.263\\
\cmark  & \xmark  &   2.383\\
\xmark  & \xmark  &  2.748 \\
\bottomrule
\end{tabular}
\end{center}
\vspace{-3.5mm}
\end{table}
\section{Conclusion}
Motion blur, in general, is regarded as an undesirable artifact. However, it contains motion information which can be processed into a more interpretable form. In this work, for the first time, we tackle the problem of estimating optical flow from a single motion-blurred image in a data-driven manner. We propose a novel and intuitive framework for the task, and successfully train it using a transfer learning approach. We show the effectiveness and generalizability of our method through a large set of experiments on synthetic and real motion-blur datasets. We also carry out in-depth analysis of our model in comparison to related approaches, guiding that naively deploying sequence restoration methods followed by standard optical flow estimation fails on this problem. The applicability of our work is also demonstrated on motion deblurring and segmentation tasks. Overall, our approach introduces a new interesting perspective on motion blur in connection with future applications such as motion deblurring, temporal super resolution and video sequence restoration from a motion-blurred image.
\vspace{-2mm}
\paragraph{Acknowledgement.}
This work was supported by NAVER LABS Corporation [SSIM: Semantic \& scalable indoor mapping].
\bibliography{egbib}
\end{document}